\documentclass[lettersize,journal]{IEEEtran}

\usepackage{amsmath,amsfonts}
\usepackage{algorithmic}
\usepackage{algorithm}
\usepackage{array}
\usepackage[caption=false,font=normalsize,labelfont=sf,textfont=sf]{subfig}
\usepackage{textcomp}
\usepackage{stfloats}
\usepackage{url}
\usepackage{verbatim}
\usepackage{graphicx}
\usepackage{amssymb}
\usepackage{cite}

\begin{document}

\title{D2CFR: Minimize Counterfactual Regret with Deep Dueling Neural Network}

\author{Huale Li, Xuan Wang, Zengyue Guo, Jiajia Zhang, Shuhan Qi
	\thanks{The corresponding author is Shuhan Qi (e-mail: shuhanqi@cs.hitsz.edu.cn)}
	\thanks{The authors are with the School of computer science and technology, Harbin institute of technology, Shenzhen 518055, China }}

\markboth{Journal of IEEE,  Oct.~2021}%
{Shell \MakeLowercase{\textit{et al.}}: A Sample Article Using IEEEtran.cls for IEEE Journals}


\maketitle

\begin{abstract}
\textit{Counterfactual Regret Minimization (CFR)} is a popular method for finding approximate Nash equilibrium in two-player zero-sum games with imperfect information. Solving large-scale games with \textit{CFR} needs a combination of abstraction techniques and certain expert knowledge, which limits its scalability. Recent neural-based CFR methods alleviate the requirement for abstraction and expert knowledge by training an efficient network to obtain the counterfactual regret directly without abstraction. However, these methods only consider the estimation of the regret value for each action, but ignore to evaluate the value of states, which is significant for decision making. In this paper, we introduces \textit{deep dueling counterfactual regret minimization (D2CFR)}, which focuses on the state value estimation by adopting a novel value network with dueling structure. Moreover, a rectification module based on a time-shifted Monte Carlo simulation is designed to rectify the inaccurate state value estimation. Extensive experimental results are conducted to show that \textit{D2CFR} converges faster and outperforms comparison methods on test games.
\end{abstract}

\begin{IEEEkeywords}
Imperfect-information games, Counterfactual regret minimization, Nash equilibrium, neural network.
\end{IEEEkeywords}

\section{Introduction}
\IEEEPARstart{I}{n} recent years, the research of imperfect-information game (IIG) has attracted more and more attention. Due to private information unobservable to other players, the IIG is usually considered to be a more complex problem than perfect information game (PIG) \cite{samuel1959some,myerson1997game}. A typical goal in IIG is to approximate an equilibrium strategy in which all players' strategies are optimal\cite{DengN19,ZhangYYL19,LiQMZK21}. Generally, the solution of two-player IIG is to find its Nash equilibrium \cite{osborne1994course,nash1951non}. Counterfactual regret minimization (CFR) is a classical method to compute Nash equilibrium in two-player IIGs \cite{zinkevich2008regret}. Recently, the latest variation of CFR and its development systems \cite{bowling2015heads,moravvcik2017deepstack,Brown2017Superhuman,schmid2019variance,brown2019superhuman}, such as DeepStack, Libratus and Pluribus, have achieved great success in the field of IIGs.

Nevertheless, there is still an inevitable problem: the scale of problems solved by CFR is not large enough with at most $10^{18}$ states, but heads-up no-limit Texas hold'em already includes almost $10^{161}$ states. It is important to note that recent successful applications of CFR all apply abstraction techniques. In particular, the original game has to be abstracted first, then the abstracted game is solved by CFR-based methods. As a result, this kind of method always requires certain domain knowledge in designing the abstraction method, which increases additional difficulty of solving large-scale games. Moreover, such abstraction will lead to the information loss, which further injures the final result.

Recently, some CFR-based methods apply neural networks to speedup the solution, which avoid the information loss caused by abstraction. DeepCFR \cite{brown2019deep} obviates the need for abstraction by using deep neural networks to approximate the behavior of CFR in the full game. DeepCFR also proves that the convergence result is  an $\epsilon$-Nash equilibrium in two-player zero-sum IIGs. Concurrent work has also investigated a similar combination of deep learning with CFR in Double Neural CFR \cite{li2018double}. This method may not be theoretically sound and it considers only small games \cite{brown2019deep}. Single DeepCFR (SD-CFR) \cite{steinberger2019single} is a variant of DeepCFR, which only applies one neural network instead of training an additional network to approximate the weighted average strategy. In addition, Neural fictitious self-play (NFSP) \cite{heinrich2015fictitious,2016Deep} is a framework to find Nash equilibrium in IIGs, which combines neural network\cite{rowley1998neural} and fictitious play \cite{Shapley1996Fictitious,lambert2005fictitious,shamma2005dynamic} to fit an average response strategy approaching Nash equilibrium. However, this method is very difficult to train in large-scale games.

Although there have been methods combining CFR with neural networks, these methods only try to fit the regret value of actions directly and ignore evaluating the importance of states. The regret value includes two essential elements: state value and action-state value. In many cases, the estimation of the state value is much more important than that of the action-state value. Therefore, in this paper, we decouple the state value and action-state value from the regret value, and apply Monte Carlo to rectify the inaccurate estimation of state value. Specifically, we introduce deep dueling CFR (D2CFR), an improved variant of DeepCFR, to solve large-scale IIGs. In our method D2CFR, a novel value network with dueling structure is adopted, whose key insight is to emphasize on the accurate evaluation of state value. Also, a rectification module based on the time-shifted Monte Carlo simulation is designed to rectify the state value estimation in the early stage of training, which speedups the convergence of the value network. We summarize our contributions as follows: 
\begin{itemize}
	\item [1)]  We present an improved variant of DeepCFR, which we call deep dueling CFR (D2CFR). The D2CFR studies on finding approximate Nash equilibrium in two-player large-scale IIGs without any abstraction, which obviates the requirement for expert knowledge. 
	\item [2)]  We propose a novel value network with dueling architecture, which aims to decouple the state value estimation and action-state value estimation. In this way, accurate evaluation of the state value will be obtained.
	\item [3)]  We design a rectification module composed of the value network and Monte Carlo simulation, which can further rectify the inaccuracy of estimation of state in the early stage.
	\item [4)]  Extensive experimental results show that D2CFR not only converges faster, but also achieves strong performance compared with DeepCFR on test games.
\end{itemize}

The rest of this paper is organized as follows. Section \ref{background} introduces the model of extensive-form game, Nash Equilibrium, CFR and MCCFR. The details of our method is described in Section 3. In Section 4, the theoretical analysis of the proposed method is given. Section 5 depicts the detail of extended experiments. Finally, we make a conclusion for this paper.

\section{Background}\label{background}
\subsection{Extensive-form Game}\label{Extensive-form Game}
In the field of IIGs, the extensive-form game is usually used to model sequential decision-making games \cite{osborne1994course}. Generally, a finite extensive-form IIG contains six components, represented as $\left<N,H,P,f_{c},I,u_i\right>$ \cite{Martin1994}: player $i$ represents a finite set $N$ of game players, $N=\{1,2,\dots,n\} $. Action history $h$ is a finite set $H$ of sequences, the possible histories of actions $a \in A$, $A(h)=\{a \mid(h, a) \in H\}$ are actions available after a nonterminal history $h \in H$. $Z \subseteq H$ are terminal histories. $P$ is the player function. $P(h)$ is the player taking action $a$ after history $h$. $P(h) = c$ represents that the chance determines the action after history $h$. A function $f_{c}$ that associates with every history $h$ for which $P(h)=c$ a probability measure $f_c(\cdot|h)$ on $A(h)$. The set $\mathrm{I}_{i} \in \mathcal{I}_{i}$ is an information set of player $i$. For any information set $I_i$, all nodes $h$, $h^{'}\in I_i$ are indistinguishable to player $i$. Payoff function $u_i$ defines the payoff of terminal state $z$ for each player $i$. For a zero-sum game, there is $u_1 = -u_2$. 

\subsection{Nash Equilibrium}\label{Nash Equilibrium}
Approximating Nash equilibrium has been proven to be an effective way in solving two-player IIGs. The Nash Equilibrium is a strategy profile in which no player can improve their utility by deviating from this strategy. The definition of strategy and best response will be given first before introducing Nash equilibrium \cite{nash1951non}.

In an extensive-form game, a strategy $\sigma_i(I)$ of player $i$ is a probability vector over actions on information set $I$. A set of strategies for players, $\sigma_{1}, \sigma_{2}, \ldots, \sigma_{n}$, makes up a strategy profile $\sigma$, and $\sigma_{-i}$ represents the strategy in $\sigma$ except the strategy $\sigma_i$ of player $i$. In addition, $\pi^{\sigma}(h)$ is the probability with $h$ occurs if all players make decision according to the strategy $\sigma$, and $\pi^{\sigma}(I)=\Sigma_{h \in I} \pi^{\sigma}(h)$. $\pi_{i}^{\sigma}(h)$ is the contribution of player $i$ to this probability. And formally, $\pi_{i}^{\sigma}(h)=\prod_{i \in N \cup\{c\}} \pi_{i}^{\sigma}(h)$. Accordingly, $\pi_{-i}^{\sigma}(h)$ of history $h$ is the contribution of all players (including chance player) except player $i$. A best response to $\sigma_{-i}$ is a strategy$ BR$$(\sigma_{-i})$, $B R\left(\sigma_{-i}\right)=\max _{\sigma_{i}^{\prime} \in \Sigma_{i}} u_{i}\left(\sigma_{i}^{\prime},  \sigma_{-i}\right)$, and $\Sigma_i$ represents all possible strategy profiles for player $i$.

A Nash equilibrium $\sigma^*$ is a strategy profile that each player plays a best response: $\forall i, u_{i}\left(\sigma_{i}^{*}, \sigma_{-i}^{*}\right)=\max _{\sigma_{i}^{\prime}} u_{i}\left(\sigma_{i}^{\prime}, \sigma_{-i}^{*}\right)$. Nash equilibrium has been proven to exist in all finite games and many infinite games. Due to it is difficult to compute Nash equilibrium in most cases, it is more common to compute approximate Nash equilibrium. An $\epsilon$-Nash equilibrium, $u_{i}\left(\sigma_{i}^{*}, \sigma_{-i}^{*}\right)+\varepsilon \geq \max _{\sigma_{i}^{\prime}} u_{i}\left(\sigma_{i}^{\prime}, \sigma_{-i}^{*}\right)$, is a strategy profile, in which no player can increase their utility by more than $\epsilon$ by changing their strategy.

\subsection{Counterfactual Regret Minimization}\label{Counterfactual Regret Minimization}
Counterfactual regret minimizing (CFR) is a classical method to find Nash equilibrium in the two-player zero-sum IIGs \cite{zinkevich2008regret}. It is an iterative solution method, which mainly includes two steps.

Step1: Calculate the total regret of action $a$ at the information $I$ on the iteration $T$. The total regret $R(I,a)$ can be depicted as:
\begin{equation}
	R_{i}^{T}(I, a)=\sum_{t=1}^{T} r_{i}^{t}(I, a)
\end{equation}
where $r_{i}^{t}(I, a)$ is instant regret on the iteration $t$, which is the difference between player $i's$ counterfactual value from playing action $a$ vs playing strategy $\sigma$ at information set $I$, $r_{i}^{t}(I, a)=v_{i}^{\sigma}(I, a)-v_{i}^{\sigma}(I)$. $v_{i}^{\sigma}(I)$ is counterfactual value and also represents the state value in this paper, which is the expected utility of player $i$ when the information set $I$ is reached, $v_{i}^{\sigma}(I)=\sum_{h \in I, h^{\prime} \in Z} \pi_{-i}^{\sigma}(h) \pi^{\sigma}\left(h, h^{\prime}\right) u_{i}\left(h^{\prime}\right)$. $v_{i}^{\sigma}(I,a)$ is counterfactual value of action $a$ and also represents action-state value in this paper, which is the same as counterfactual value $v_{i}^{\sigma}(I)$ except that the player $i$ selects action $a$ all the time at information set $I$. In addition, the positive regret is only considered in most cases, $R_{i}^{T,+}(I, a)=\max \left(R_{i}^{T}(I, a), 0\right)$.

Step2: Update the strategy $\sigma_{i}^{T+1}(I,a)$ of next iteration $T+1$. Regret matching (RM) algorithm \cite{foster1999regret} is used to update the strategy on each iteration. Formally, the strategy on the iteration $T+1$ can be calculated as follows:
\begin{equation}
	\sigma_{i}^{T+1}(I,a)=\left\{\begin{array}{ll}
		\frac{R_{i}^{T,+}(I, a)}{\sum_{a' \in A(I)} R_{i}^{T,+}\left(I, a^{\prime}\right)},&\sum_{a \in A(\mathrm{I})} R_{i}^{T,+}(I, a)>0 \\
		\frac{1}{|A(I)|}, & \text { otherwise }
	\end{array}\right.
\end{equation}

If a player plays according to CFR on each iteration, then $R_{i}^{T} \leq \sum_{I \in \mathcal{I}_{i}} R^{T}(I)$. So, as $T \rightarrow \infty, \frac{R_{i}^{T}}{T} \rightarrow 0$. Moreover, the average strategy $\left\langle\bar{\sigma}_{1}^{T}, \bar{\sigma}_{2}^{T}\right\rangle$ form a $2\epsilon$-Nash equilibrium, if the average total regret of both player satisfies $\frac{R_{i}^{T}}{T} \leq \epsilon$ in two-player zero-sum games. And the average strategy for information set $I$ on iteration $T$ is $\bar{\sigma}_{i}^{T}(I)=\frac{\sum_{t=1}^{T} \pi_{i}^{\sigma}(I) \sigma^{t}(I)}{\sum_{t=1}^{T} \pi_{i}^{\sigma}(I)}$.

\subsection{Monte Carlo CFR}\label{MCCFR}
Vanilla CFR needs to traverse the full game tree of games, which limits its application in large games. Towards this problem, Monte Carlo (MCCFR) \cite{lanctot2009monte} expands the solution scale of the vanilla CFR, which only needs to traverse a portion of the game tree. In this paper, we also adopt the MCCFR as the main scheme.  Instead of using counterfactual value in the vanilla CFR, the MCCFR utilized sampled counterfactual value, which is defined as follows:
\begin{equation}
	\tilde{v}_{i}(\sigma, I \mid j)=\sum_{z \in Q_{j} \cap Z_{I}} \frac{1}{q(Z)} u_{i}(z) \pi_{-i}^{\sigma}\left(z[I] \pi_{\sigma}(z[I], z)\right)
\end{equation}
where $\mathrm{Q}=\left\{Q_{1}, \cdots, Q_{r}\right\}$ is a set of subsets of $Z$, and one of these subsets can be called a block. $q_j$ is the probability when block $Q_j$ is considered for current iteration, and $\sum_{j=1}^{r} q_{j}=1, q_{j}>0$. $\mathrm{q}(\mathrm{z})=\sum_{j: z \in Q_{j}} q_{j}$. And the sampled immediate counterfactual regret $\tilde{r}(I, a)$ of action $a$ is:
\begin{equation}
	\tilde{r}(I, a)=\tilde{v}_{i}\left(\sigma_{I \rightarrow a}^{t}, I\right)-\tilde{v}_{i}\left(\sigma^{t}, I\right)
\end{equation}

The same with CFR, the strategy of next iteration can be calculated with the RM. Moreover, \cite{lanctot2009monte} has proved that the counterfactual value of MCCFR is the same with the normal CFR on the expectation value. 

MCCFR includes two kinds of sampling methods: outcome-sampling (OS) and external-sampling (ES). In ES-MCCFR, the action is sampled when it comes from the opponent and chance player. The sampled counterfactual value of every visited information set can be calculated as follows:
\begin{equation}
	\tilde{r}(I, a)=(1-\sigma(a \mid z[I])) \sum_{z \in Q \cap Z_{I}} u_{i}(z) \pi_{i}^{\sigma}(z[I] a, z)
\end{equation}

Lanctot et al. \cite{lanctot2009monte} gave a proof that the average strategy solved by ES-MCCFR converges Nash equilibrium as the iteration increases. And its average overall regret is bounded by $R_{i}^{T} \leq\left(1+\frac{\sqrt{2}}{\sqrt{p}}\right) \Delta_{u, i} M_{i} \sqrt{\left|A_{i}\right|} / \sqrt{T}$.

\subsection{A Conventional solving framework with CFR}\label{appea}
\begin{figure}[h]
	\includegraphics[width=3in]{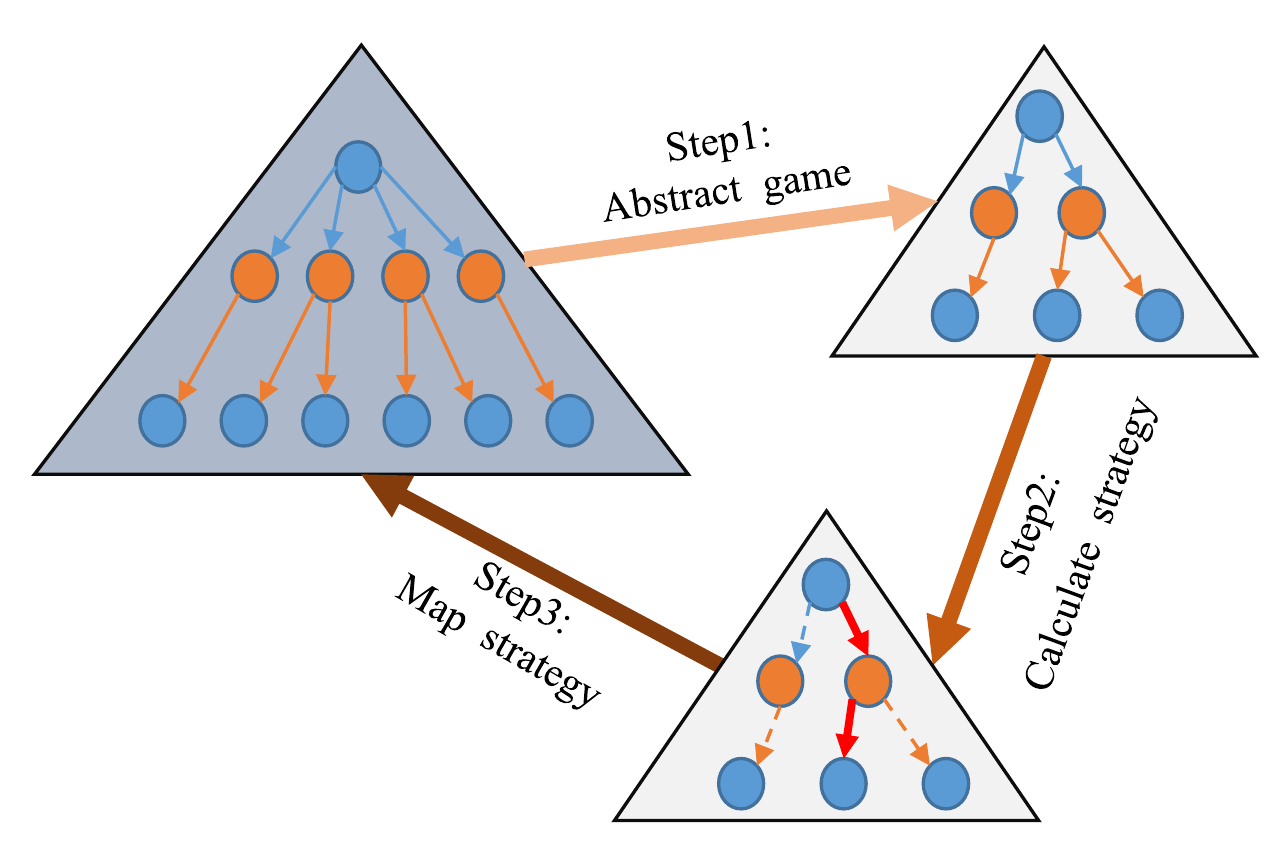}
	\caption{Conventional framework of solving strategy with CFR. The big triangle in the upper left corner is the original game, and inside the triangle is the game tree that is representation of the original game. The small triangle in the upper right corner is the game after abstraction, and the bottom triangle is equivalent to the abstract game. The final strategy can be obtained with three steps: abstract game, calculate strategy, map strategy.}
	\label{traframework}
\end{figure}

Here we review the conventional framework of CFR for solving IIGs. Restricted by solution scale of CFR, the game needs to be reduced to the scale that CFR can be solved. The conventional framework of CFR is showed in Fig.\ref{traframework}. Generally, it includes three steps to obtain the final strategy. Firstly, abstract the original game\cite{shi2000abstraction,gilpin2007potential,gilpin2006competitive,gilpin2007better}. Then, apply CFR to solve the strategy of the abstracted game. Finally, map the solved strategy back to the original game\cite{ganzfried2013action}. However, it can be clearly found that whether abstracting the original game or mapping strategy will bring certain errors to final results. 

\section{Our method}\label{our method}
\subsection{An overview of the framework}\label{overview of framework}
\begin{figure}[t]
	\includegraphics[width=3.3in]{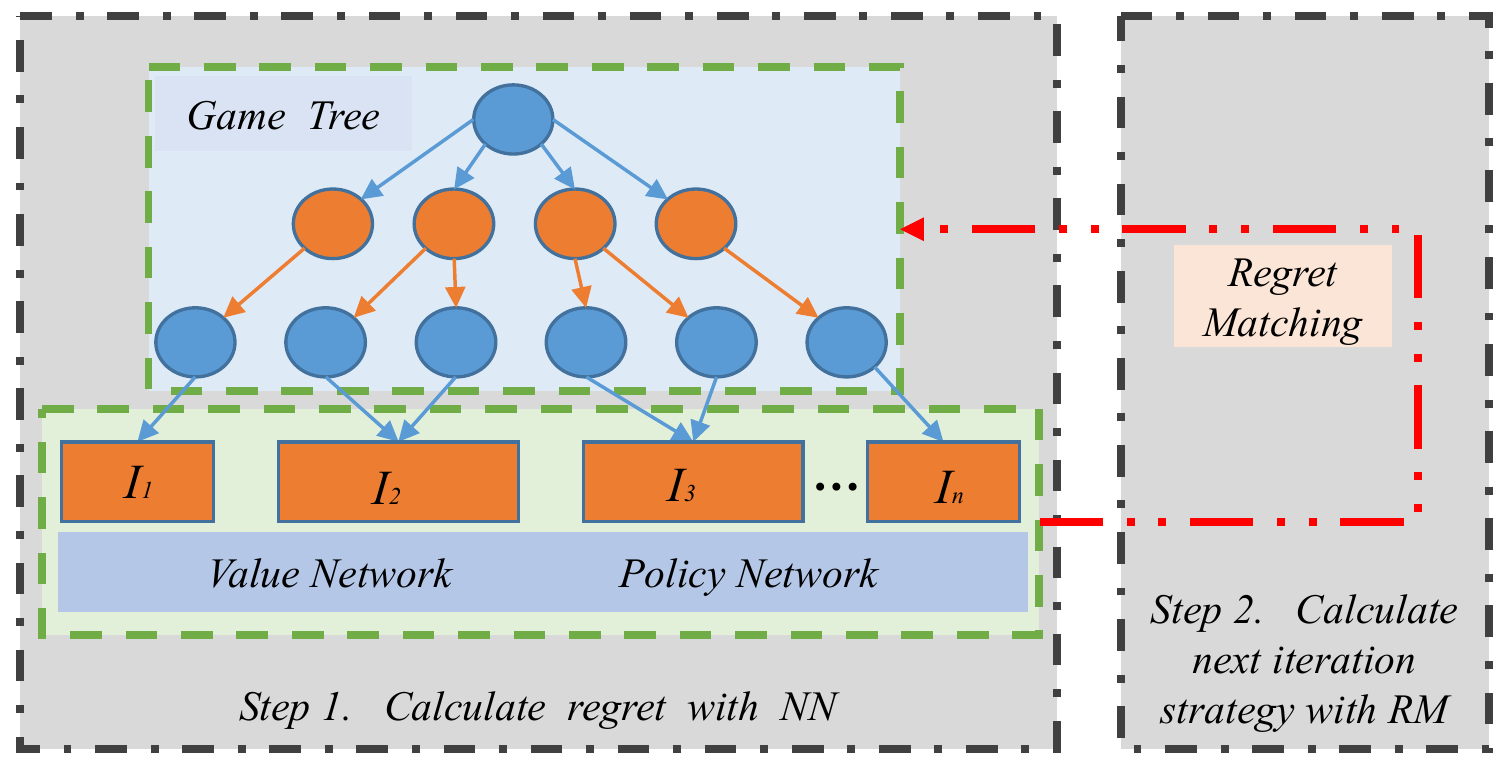}
	\caption{The framework of our approach D2CFR. Our approach is divided into two steps: calculate the regret value and calculate the strategy of next iteration. The diagram here takes a two-player game as an example, where the blue circle represents game player 1 and the orange circle represents game player 2. The arrow between the two circles indicates the action. The orange box represents information sets $I_1, I_2, \dots, I_n$.}
	\label{CFRNN}
\end{figure}

Now we give an overview of the proposed D2CFR. As shown in Fig.\ref{CFRNN}, each iteration of D2CFR can be divided into two steps. For step 1, the instant regret value is fitted by the value network in each iteration, and for step 2, the strategy of next iteration is calculated with the RM. By repeating these two steps, the strategy will approximate to a Nash equilibrium strategy, as long as the number of iterations is large enough. Here the D2CFR conducts IIG solving in an end-to-end way and without any abstraction, which largely alleviates the requirement for expert knowledge. 

Different from CFR, the regret value in D2CFR is approximated by neural network instead of fully expanding the game tree. Similar with the DeepCFR, in progress of iteration, the value network and strategy network of D2CFR will also be trained continuously. That is, the value network is used to fit the instant regret value, while the policy network is an approximation of the average strategy. It is worth noting that, compared with DeepCFR, the proposed D2CFR decouples the state value from the regret value calculation and tries to learn these two values jointly. Moreover, the D2CFR introduces a time-shifted Monte-Carlo Simulation to rectify the state value estimation.

\subsection{Description of our method}\label{Description of our method}
In this section, the value network, rectification module, policy network and loss function of D2CFR will be described in detail respectively.

\textbf{Value network:} As described in Eqs.1-2, one of precondition of updating the strategy by RM is to obtain the instant regret value of each action in information sets. In the proposed D2CFR, the instant regret value is estimated by the value network. We adopt an end-to-end way to obtain the strategy, instead of traversing the whole game tree completely in CFR. This is the essential difference between our method and conventional CFR. Here the end-to-end means from game state to game strategy. To be specific, when a given information set is used as input, the instant regret value of each action is directly outputed through the value network. Here each action refers to all legal actions at the current information set.

\begin{figure*}
	\includegraphics[width=7in]{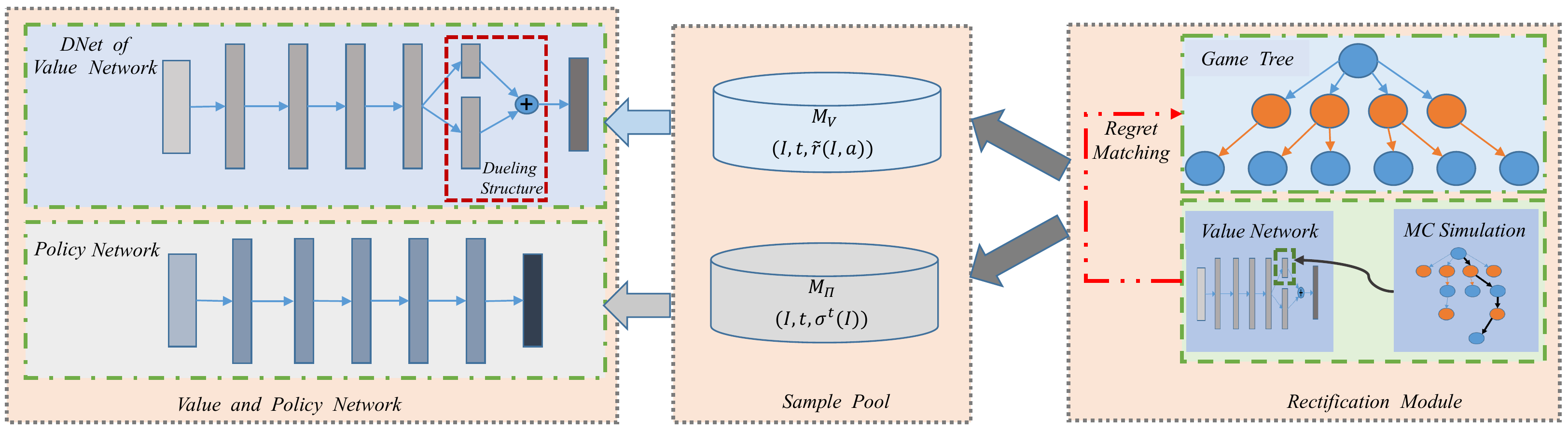}
	\caption{Training neural networks with rectification module. On each iteration $t$, D2CFR conducts $K$ traversals of the game tree partially, with the path of the traversal determined by ES-MCCFR. When encountered the information set $I$, it plays a strategy computed by the RM on output of the value network. Samples of training value network are collected through the rectification module. }
	\label{NN}
\end{figure*}

Compared with the fully connected network structure in DeepCFR, the value network in D2CFR adopt a novel dueling network structure (called DNet), which aims to decouples counterfactual value (i.e., state value) and action counterfactual value (i.e., action-state value) from the instant regret estimation. As shown in Fig.\ref{NN}, the dueling structure includes two sub-layers: the shorter one denotes the sub-layer used to explicitly estimate the mean counterfacutal value of information set, while the longer one represents the counterfactual value of an action for the information set. The two sub-layer shares a common feature learning module.

In the proposed DNet, the dueling network is not composed by simply dividing the fully connection layer into two sub-layers. In order to make the counterfactual value estimation more accurately, multiply loss function have been utilized to learn the counterfactural value and instant regret value simultaneously. The key insight here is that estimating the counterfactual values accurately is more important then the action counterfactual value. Since for many states, estimating the counterfactual value of all action is unnecessary. For example, in the perflop round of Leduc game, the action call and raise have almost the same effect on the probability of winning, when private and public card are both Ace. Thus, in many cases, obtain the exact value of each action is unnecessary, and the reasonable decision can be made as long as the state value is known. In this way, evaluate the counterfactual value, which reflects the value of state, is very important for instant regret value calculation. 

\textbf{Rectification module:} As mentioned above, the counterfactual value is valuable for instant regret values calculation. However, its estimation is a difficult task, especially in the early stage of iteration. This is because CFR is an iterative method, and making the strategy converge is a slow progress. In other words, in the early stage of iteration, the ground truth for training the network is totally inaccurate. This problem greatly limits the speed of strategy learning.

In this paper, we designed a rectification module to improve the accuracy of the state estimation by introducing Monte Carlo (MC)\cite{zio2013monte}. The rectification module consists of two parts: value network and Monte Carlo (MC) \cite{2012A}. In the rectification module, to estimate the counterfactual value of current state, a time-shift weighted combination of MC simulation and value network is adopted. To be specific, the counterfactual value and the action counterfactual value are output respectively in the penultimate network in the DNet, as $r_{i}^{t}(I, a)=v_{i}^{\sigma}(I, a)-v_{i}^{\sigma}(I)$ described. In our rectification module, the counterfactual value $v_{i}^{\sigma}(I)$ in the equation comes from two parts: one is still from the DNet, represented with $v_{i,NN}^{\sigma}(I)$; the other one is from the MC, represented with $v_{i,MC}^{\sigma}(I)$. Here the $v_{i,MC}^{\sigma}(I)$ is the value from MC simulation. And finally the $v_{i}^{\sigma}(I)$ in original DNet is replaced with the combination of  $v_{i,NN}^{\sigma}(I)$ and $v_{i,MC}^{\sigma}(I)$. It can be formally described with:
\begin{equation}
	v_{i}^{\sigma}(I)=\alpha v_{i,NN}^{\sigma}(I)+\beta v_{i,MC}^{\sigma}(I)
\end{equation}
here $\alpha+\beta=1$. It is worth noting that $\alpha$ and $\beta$ are varying with iterations, $\alpha=0.01+t/(t+1)$, $\beta=0.99-t/(t+1)$. This means that in the early iterative training, the model relies on the MC simulation to update the model. With the progress of training, the model increasingly believes in the estimation from the value network itself.

\textbf{Policy network:} Similar with the DeepCFR, we propose a policy network to learn the average strategy for final decision, which is shown in Fig.\ref{NN}. Actually, it is a simple but effective method for strategy learning. For CFR-based method, the average strategy obtained by iterative learning will approach to an Nash equilibrium strategy. As described in DeepCFR, using a neural network to approximate the average strategy will eventually leads to a good policy network. Since we do not adopt the average strategy in the training, there is no need to consider the large approximation error in the early stage. Thus it is reasonable to approximate average strategy by full connection network.

\textbf{Loss function:} Compared with the loss function of mean square error (MSE) in DeepCFR \cite{brown2019deep}, our method adopts a loss function combined MSE and KL-divergence\cite{bu2018estimation}. The formally definition is as follows:
\begin{equation}
	\begin{split}
		\mathcal{L}(\theta_\Pi)=\mathbb{E}_{(I,t,\sigma^{t'} )  \sim  M_\Pi } \bigg[t'\sum_{a}(\sigma^{t'}(a)-\Pi(I,a \mid \theta_\Pi))^2 - \\
		\gamma KL[\Pi^t_{\theta_\Pi}(\cdot \mid I), \frac{1}{c}\sum_{t-c}^{t}{\Pi^t_{\theta_\Pi}(\cdot \mid I)}] \bigg]
	\end{split}
\end{equation}
where $\gamma, c$ are constants and $t, t'$ are the number of iteration. 

The overall algorithms of our method D2CFR are shown in algorithm \ref{algorithm1} and algorithm \ref{algorithm2}. It can be found that D2CFR traverses the game tree several times by using ES-MCCFR for each player on each iteration. In the procedure of traversing the game tree, the RM calculates the strategy of next iteration through regret values, which are fitted by a rectification module. Besides, the value network and strategy network are constantly trained and optimized with samples that are collected by traversing the game tree. Finally, the average strategy is approximated by the policy network, which is able to approach the approximate Nash equilibrium. 

\begin{algorithm*}[t]
	\caption{Deep Dueling Counterfactual Regret Minimization} \label{algorithm1}
	\hspace{0.5cm} {\bf Input:} 
	the game $G$, ES-MCCFR iteration number $T$, traverse number $K$, constants $c$, $\alpha$, $\beta$ and $\gamma$, regret value network parameters $\theta$ for each player, regret value memories $M_{V,1}, M_{V,2}$, strategy memory $M_\Pi$, Monte Carlo times $N$.  
	
	\hspace{0.5cm} {\bf Output:}  $\theta_\Pi.$
	\begin{algorithmic}[1]
		\STATE \hspace{0.5cm}Env = $G$.  \\
		\STATE \hspace{0.5cm}Initialize each player's value network $r_{NN}\left(I,a\mid\theta_{i}\right)$ with paremeters $\theta_i$. \\
		\STATE \hspace{0.5cm}Initialize reservoir-sampled regret value memories $M_{V,1}, M_{V,2}$ and strategy memory $M_\Pi$. 
		\FOR{ES-MCCFR iteration $t=1$ to $T$} 
		\FOR{each player $i$}
		\FOR{traversal $k=1$ to $K$}
		\STATE \hspace{0.5cm}Traverse$\left(\emptyset, i, \theta_{1}, \theta_{2}, \mathcal{M}_{V, i}, \mathcal{M}_{\Pi}, t, N, \alpha, \beta \right)$  \quad $\vartriangleright$ Collect data from the game $G$ traversal with ES-MCCFR 
		\ENDFOR
		\STATE \hspace{0.5cm}Train $\theta_i$ from scratch with loss $\mathcal{L}(\theta_i)=\mathbb{E}_{(I,t,\tilde{r}^{t'} )  \sim  M_{V,i} } [t'\sum_{a}(\tilde{r}^{t'}(a)-r_{NN}(I,a \mid\theta_i))^2]$
		\ENDFOR
		\ENDFOR
		\STATE \hspace{0.5cm} Train $\theta_\Pi$ with loss $\mathcal{L}(\theta_\Pi)=\mathbb{E}_{(I,t,\sigma^{t'} )  \sim  M_\Pi } \bigg[t'\sum_{a}(\sigma^{t'}(a)-\Pi(I,a \mid \theta_\Pi))^2 - \gamma KL[\Pi^t_{\theta_\Pi}(\cdot \mid I), \frac{1}{c}\sum_{t-c}^{t}{\Pi^t_{\theta_\Pi}(\cdot \mid I)}] \bigg]$
		\RETURN $\theta_\Pi$
	\end{algorithmic}
\end{algorithm*}

\begin{algorithm*}[t]
	\caption{Traverse the game with ES-MCCFR} \label{algorithm2}
	\textbf{function} Traverse$\left(h, i, \theta_{1}, \theta_{2}, \mathcal{M}_{V, i}, \mathcal{M}_{\Pi}, t, N, \alpha, \beta \right)$  \\
	\hspace*{0.02in} {\bf Input:} 
	history $h$, traversal player $i$, regret value network parameters $\theta$ for each player, regret value memory $M_V$ for each player $i$, strategy memory $M_\Pi$, ES-MCCFR iteration $t$, Monte Carlo times $N$, constants $\alpha$ and $\beta$. \\
	\begin{algorithmic}[1]
		\IF{$h$ is a terminal node} 
		\STATE \hspace{0.5cm} \textbf{return} the payoff of the player $i$
		\ELSIF{$h$ is a chance node}
		\STATE \hspace{0.5cm} $a \sim \sigma(h)$ 
		\STATE \hspace{0.5cm}	\textbf{return} Traverse$\left(h\cdot a, i, \theta_{1}, \theta_{2}, \mathcal{M}_{V, i}, \mathcal{M}_{\Pi}, t, N, \alpha, \beta \right)$
		\ELSE
		{  \IF{it's the traverser's turn to act}
		\STATE \hspace{0.5cm}Compute strategy $\sigma^t(I)$ from predicted regret values $r_{NN}(I(h),a\mid\theta_i)$ of rectification module by using the RM.  $r_{NN}(I(h),a\mid\theta_i)=V(I(h),a\mid \theta_i)-V(I(h)) $
		\STATE \hspace{0.5cm}The predicted counterfactual value of rectification module $V(I(h)) = \alpha V_{NN}(I(h)\mid \theta_{i}) + \beta V_{MC}(I(h)) $.
		\STATE \hspace{0.5cm} $V_{NN}(I(h))$ is from the predicted value of value network,   $V_{MC}(I(h))$ is from the value of $N$ simulations in the current information set with MC.
		\FOR{$a\in A(h)$} 
		\STATE \hspace{0.5cm} $v(a)\leftarrow$ Traverse$\left(h\cdot a, i, \theta_{1}, \theta_{2}, \mathcal{M}_{V, i}, \mathcal{M}_{\Pi}, t, N, \alpha, \beta \right)$   \qquad \qquad \qquad $\vartriangleright$ Traverse each action
		\STATE \hspace{0.5cm} $\tilde{r}(I,a) \leftarrow v(a)-\sum_{a'\in A(h)} \sigma(I,a')\cdot v(a')       $  \qquad \qquad \quad $\vartriangleright$ Compute regret values
		\ENDFOR   
		\STATE \hspace{0.5cm} Insert the information set and its action regret values $(I,t,\tilde{r}^t(I))$ into the regret value memory $M_V$
		\ELSE   
		{
		\STATE \hspace{0.5cm} Compute strategy $\sigma^t(I)$ from predicted regret values $r_{NN}(I(h),a\mid \theta_{-i})$ of rectification module by using the RM. $r_{NN}(I(h),a\mid\theta_{-i})=V(I(h),a\mid \theta_{-i})-V(I(h)) $
		\STATE \hspace{0.5cm} The predicted counterfactual value of rectification module $V(I(h)) = \alpha V_{NN}(I(h)\mid \theta_{-i}) + \beta V_{MC}(I(h)) $.
		\STATE \hspace{0.5cm} Insert the information set and its action probality $(I, t, \sigma^t(I))$ into the strategy memory $M_\Pi$.
		\STATE \hspace{0.5cm} Sample an action $a$ from the probability distribution $\sigma^t(I)$.
		\STATE \hspace{0.5cm} \RETURN Traverse$\left(h\cdot a, i, \theta_{1}, \theta_{2}, \mathcal{M}_{V, i}, \mathcal{M}_{\Pi}, t, N, \alpha, \beta \right)$
     	}
      \ENDIF
	    }
		\ENDIF
	\end{algorithmic}
\end{algorithm*}

\section{Theoretical analysis}\label{Theoretical analysis for DDCFR}

In DeepCFR \cite{brown2019deep}, it proves that with probability $1-\rho$ total regret at iteration $T$ is bounded by $R^T_p \leqslant (1+\frac{\sqrt{2}}{\sqrt{\rho K}})\Delta |\mathcal{I}_p| \sqrt{|A|} \sqrt{T}+4T\mathcal{I}_p\sqrt{|A| \Delta \epsilon_\mathcal{L}}$ with probability $1-\rho$. And $\mathcal{L}_V^t-\mathcal{L}_{V^*}^t \leqslant \epsilon_\mathcal{L}$, where $\mathcal{L}_V^t$ is the average mse loss for $V_p(I,a|\theta^t)$ on a sample in $M_{V,p}$ at iteration $t$, $\mathcal{L}_{V^*}^t$ is the minimum loss achievable for any function $V$. In our method D2CFR, $\mathcal{L}_V^t$ is different since the function $V$ is different from DeepCFR's function $V$. Because samples in the value memory come from a rectification module, which composed of value network and MC. The function in D2CFR is $r_{NN}(I(h),a\mid\theta_i)=V(I(h),a\mid \theta_i)-V(I(h)) $, and $V(I(h)|\theta_i) = \alpha V_{NN}(I(h)|\theta_i) + \beta V_{MC}(I(h)) $, where $\alpha+\beta=1$. Here $V_{MC}$ obtained by MC is a finite constant. Thus the function satisfies $V(I(h),a|\theta_i) \leq \alpha V_{NN}(I(h),a|\theta_i) + \omega \beta V_{NN}(I(h),a| \theta_i) $, where $\omega$ is a finite constant. Furthermore, the function can be rewrited with $V(I(h),a|\theta_i) \leq (\alpha+\omega \beta) V_{NN}(I(h),a|\theta_i)$, where $\alpha+\omega \beta$ is a finite constant. At present, our function has been formally consistent with DeepCFR's function $V$. In addition, in D2CFR, $\mathcal{L}_V^t-\mathcal{L}_{V^*}^t \leqslant \omega \epsilon_\mathcal{L}$. In this case, we obtain the conclusion that the regret can be bounded by $R^T_p \leqslant (1+\frac{\sqrt{2}}{\sqrt{\rho K}})\Delta |\mathcal{I}_p| \sqrt{|A|} \sqrt{T}+4T\mathcal{I}_p\sqrt{|A| \Delta \omega \epsilon_\mathcal{L}}$ with probability $1-\rho$, as the other conditions are the same as DeepCFR. Finally, the same as DeepCFR, as $T \rightarrow \infty$, the average regret $\frac{R_i^T}{T}$ is bounded by $4\mathcal{I}_p\sqrt{\omega|A| \Delta \epsilon_\mathcal{L}}$. 

\section{Experiments}\label{experiments}
In this section, the experimental setup and experimental results are introduced. The testbed, implementation and evaluation metric will be detailed described in the experimental setup. The comparison experiments and ablation studies are conducted in the experimental results.

\subsection{Experimental setup}\label{setup}
\subsubsection{Experimental testbed}
Poker is a family of games that includes hidden information, deception, and bluffing, which has been used as a domain for testing of game-theoretic techniques in the field of IIGs \cite{bowling2015heads,moravvcik2017deepstack,Brown2017Superhuman,schmid2019variance,brown2019superhuman}. Many successful CFR-based methods and applications take poker games as the testbed to verify their effectiveness, such as DeepStack \cite{moravvcik2017deepstack}, Libratus \cite{Brown2017Superhuman}, Pluribus \cite{brown2019superhuman}. In this paper, Leduc hold'em \cite{southey2005bayes} and Heads-up No-limit Texas hold'em (HNLH) are used to test the effectiveness of our method. These two games are both two-player games. 

Game rules of Leduc: Leduc hold'em is a popular benchmark for IIGs since its size and strategic complexity. In Leduc hold'em, there is totally six cards: two each of Jack, Queen, and King. There are two rounds: preflop and flop. In the round of preflop, each player is dealt one card as private card respectively and an ante of 1 is placed in the pot. The player 1 goes first and the maximum betting number is 2 in the preflop round. And then one public card is dealt before the flop round begins. The palyer 1 goes first again and the maximum betting number is 2 in the flop round. If one of the players has a pair with the public card, that players wins. Otherwise, the player with the higher card wins.

Game rules of HNLH: HNLH is a two-player imperfect-information game. HNLH totally contains 52 cards and consists of four rounds. The four betting rounds are preflop, flop, turn and river. Three kinds of actions, fold, call and raise can be chosen by each player on a round of betting. If the acting player chooses the action fold, it means this player is out of the current game and cannot obtain any chip in this game. If the acting player chooses the action call, it means this player bets chips into the pot. The number of betting chips should be equal to the most chips that other players have contributed to the pot. If the acting player chooses the action raise, it means this player can add more chips to the pot. And the number of raising chips should be more than any other player raised so far. Besides, there is no limit to the number of times a player can raise. The player can choose how much to raise and the subsequent raise on each round should be at least as large as the previous raising chips.

At the beginning of the preflop round, two cards are dealt to each player from a standard 52-deck respectively. And it should be noted that these two cards are private cards for each player, which are unobservable to each other. Three public cards are dealt in the flop round and a public card is dealt in the last two round respectively. Here the public card represents this card can be observable to each player. The player will be the winner and obtain all pot chips when this player is the only remaining player in the game. Otherwise, the player with five best cards that consists of two private cards of the player and three public cards from five public cards, wins the pot. In the case of a tie, the pot will be splitted equally to winning players.

\subsubsection{Implementation detail}
Our experiments are conducted on the platform OpenSpiel \cite{lanctot2019openspiel}, which is a collection of environments and algorithms for research in games. Where not otherwise noted, the comparison algorithm DeepCFR and NFSP both are trained completely according to the algorithm provided by the OpenSpiel. For the SD-CFR, we reproduced it according to the original paper \cite{steinberger2019single}. All parameters are set completely according to the original paper. 

For our D2CFR, we set hyperparameters as follow. For the DNet, it includes seven layers, the information sets are taken as input and outputs the regret value of each action. The policy network has seven fully connected layers and outputs probability of each legal actions. The batch size is 200. The parameters are updated by Adam optimizer\cite{kingma2014adam} with a learning rate 0.001. The memory capacity is 100,000. The total number of iteration $T$ is 1000, which is enough for all methods. For the times of MC in rectification module, the times $N$ is 800 and $\alpha=0.01+t/(t+1)$, $\beta=0.99-t/(t+1)$. The parameters $c=10$, $\gamma=0.05$. In addition, all experiments are conducted on four Xeon(R) CPUs of E5-2640 with 10 cores $@$2.40GHz, and one Tesla P100 GPU with 16G memory.

\subsubsection{Evaluation metric}\label{metric}
In this paper, the effectiveness of our method will be evaluated with two popular metrics in this field: exploitability and head-to-head performance, just like the experimental setup in previous works \cite{Brown2017Superhuman,lanctot2019openspiel,moravvcik2017deepstack,zinkevich2008regret}.

Exploitability is a standard metric, which is used to measure the strategy in two-player IIGs. The exploitability $e(\sigma_i)$ of strategy $\sigma_i$ indicates how close $\sigma$ is to a Nash equilibrium strategy in a two-player zero-sum game. And the lower the exploitability, the better the strategy. The exploitability $e(\sigma_i)$ is formally defined as:
\begin{equation}
	e(\sigma_i) = u_i(\sigma^*_i,BR(\sigma_i^*)) - u_i(\sigma_i,BR(\sigma_i))
\end{equation}
where $BR(\sigma_i)$ is the best response to the strategy $\sigma_i$, which has been detailed introduced in background.

Considering that the exploitability only can be calculated in small-scale games, it is only used in the evaluation of Leduc hold'em. For the HNLH, the head-to-head performance is measured to further verify the effectiveness of our method. In our experiments, 10,000 games are conducted for each comparison respectively.

\subsection{Experimental results}\label{results}
Two groups of experiments were conducted. The first group experiment is to verify the effectiveness of D2CFR compared with previous state-of-the-art methods NFSP \cite{2016Deep}, DeepCFR \cite{brown2019deep} and SD-CFR \cite{steinberger2019single}, on the HNLH. The second group experiment is an ablation study, which analyzed the effect of each improved component (DNet and rectification module).   

\subsubsection{comparision with state-of-the-art methods}
\begin{figure}[t]
	\includegraphics[width=3in]{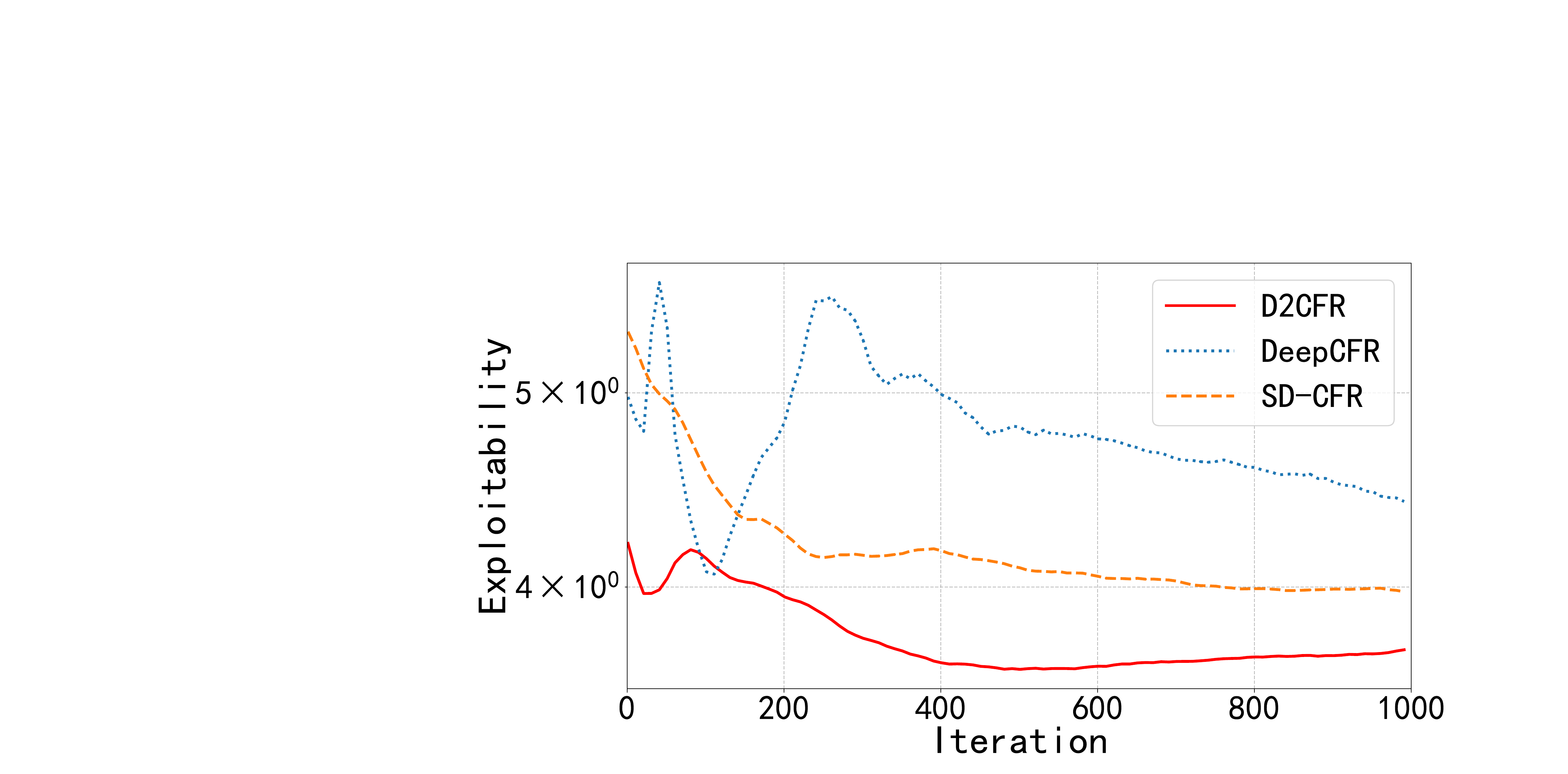}%
	\centering
	\caption{Exploitability. The Y-axis is the exploitability and the X-axis is the number of iteration. The lower , the better.}
	\label{explo}
\end{figure}

We conducted the first group experiment with three state-of-the-art methods in recent years, which are NFSP \cite{2016Deep}, DeepCFR \cite{brown2019deep}, SD-CFR \cite{steinberger2019single}. In order to show the performance of our method, we first test the exploitability on the Leduc compared with other two CFR-based methods DeepCFR and SD-CFR, as shown in Fig.\ref{explo}. Secondly, the head-to-head performance is tested on both Leduc and HNLH, as shown in Tab.\ref{tab:headhulh}.

Fig.\ref{explo} shows that D2CFR reaches a lower level of exploitability compared with DeepCFR and SD-CFR. Specifically, compared with the DeepCFR, the exploitability of D2CFR has been significantly lower than that of DeepCFR except $110th\sim120th$ iterations. This result shows that D2CFR is very effective in improving the DeepCFR. In addition, we found that except in the early $170th$ iterations, the SD-CFR outperforms the DeepCFR in exploitability. This further shows that our replication of SD-CFR is successful and the experimental result is credible. 

\begin{table}[!t]
	\caption{Head-to-head performance of D2CFR\label{tab:headhulh}}
	\centering
	\resizebox{\linewidth}{!}
	{
		\begin{tabular}{|c||c||c||c||c|}
			\hline  
			 Tested game&&DeepCFR&SD-CFR&NFSP  \\ \hline
			Leduc	& D2CFR	vs& 276.1$\pm$47.2 &40.85$\pm$ 9.2   & 219$\pm$28.33    \\ \hline
			
			HNLH& D2CFR	vs&  75.52$\pm$16.0 &64.08$\pm$ 15.7   & 119.44$\pm$27.48    \\
			\hline
		\end{tabular}
	}	
\end{table}

Tab.\ref{tab:headhulh} gives the detailed head-to-head performance on the Leduc and HULH respectively, which was measured in milli–big blinds per game ($mbb/g$), the average number of big blinds won per 1000 games. We report average winning in $mbb/g$ followed by the $95\%$ confidence interval. The D2CFR defeated DeepCFR, SD-CFR and NFSP by 276.1$\pm$47.2 $mbb/g$, 40.85$\pm$ 9.2 $mbb/g$ and 219$\pm$28.33 $mbb/g$ on the Leduc respectively. And on the HNLH, the D2CFR defeated DeepCFR, SD-CFR and NFSP by 75.52$\pm$16.0 $mbb/g$, 64.08$\pm$15.7 $mbb/g$ and 119.44$\pm$27.48 $mbb/g$ respectively. These results show that our D2CFR is obviously better than the comparison methods.

To sum up, firstly, our method shows much better performance than the DeepCFR in terms of exploitability and head-to-head performance. It shows that our improvement on the DeepCFR is very effective. Secondly, the better results of comparative experiments with SD-CFR and NFSP further verify the excellent performance of our approach D2CFR.

\subsubsection{Ablation study}
The ablation study was conducted in the third group experiment, which includes three aspects. First, the effectiveness of the DNet is evaluated. Second, the experiment is conducted to test the performance of the rectification module.  Third, the different setting values of $N$ in MC is conducted.

\begin{figure*}[t!]
	\centering
	\subfloat[Winning ]{\includegraphics[width=0.48\linewidth,height=0.25\linewidth]{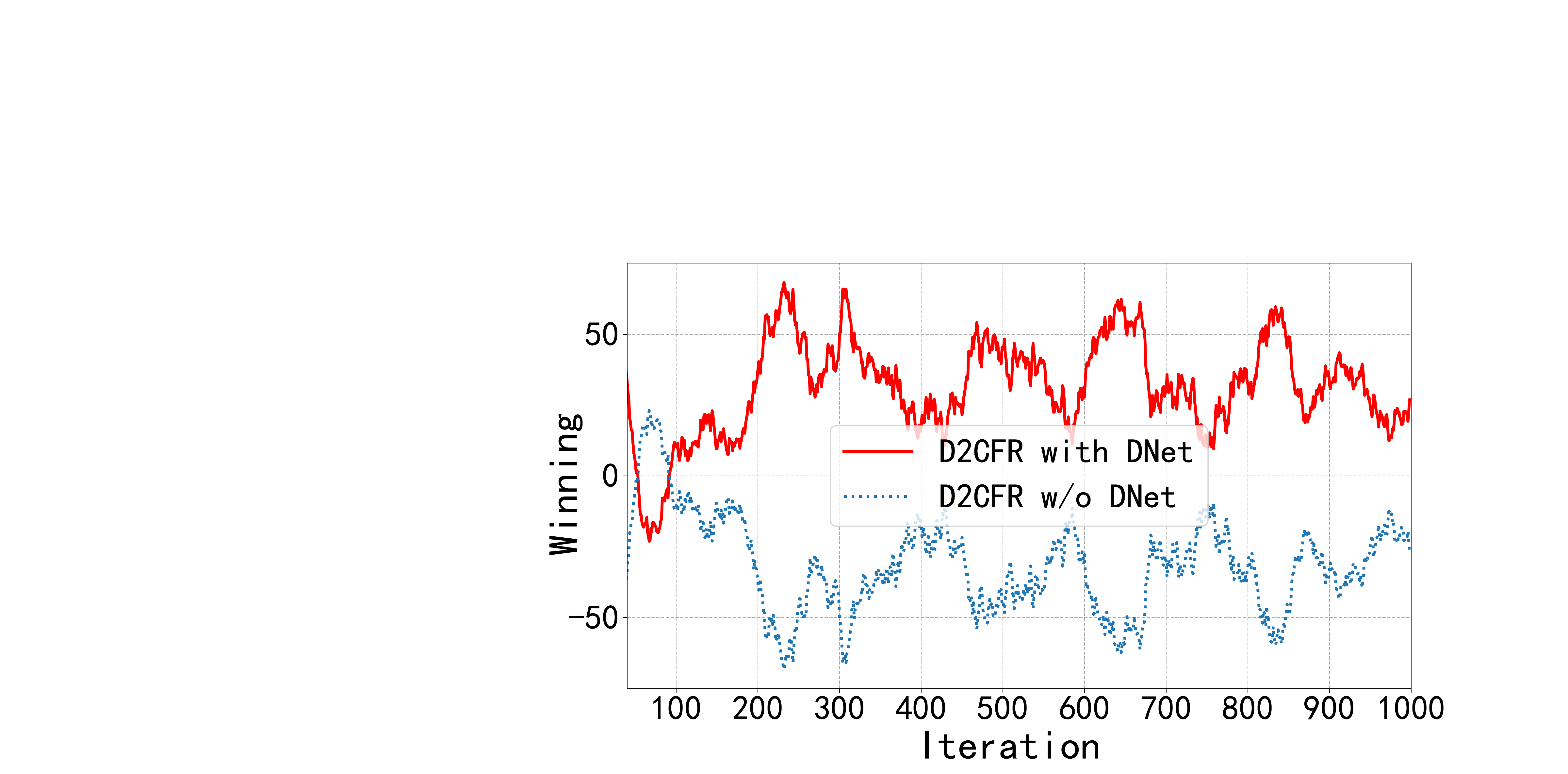}%
	}
	\hfil
	\subfloat[Policy loss ]{\includegraphics[width=0.48\linewidth,height=0.25\linewidth]{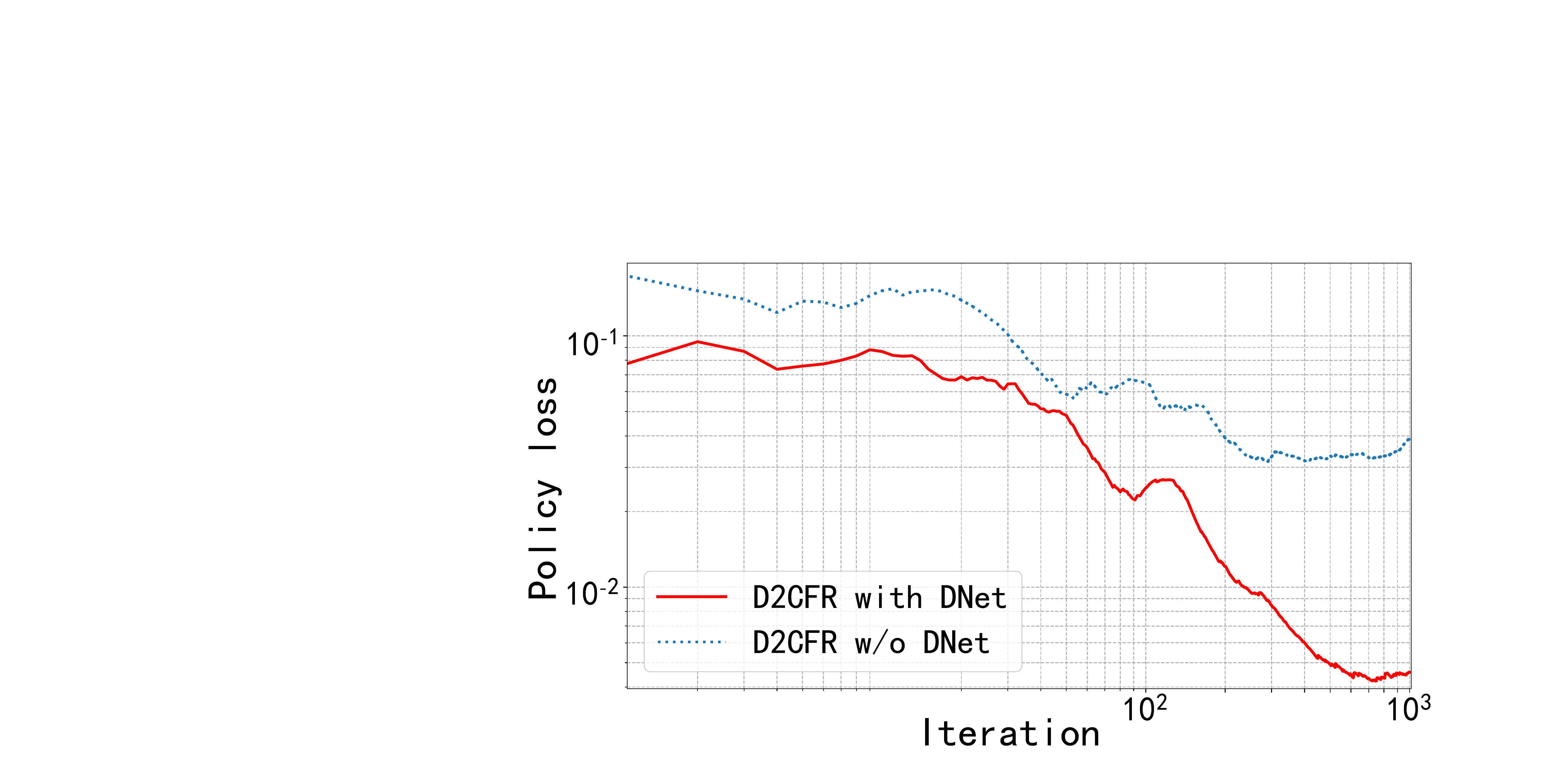}%
	}
	\caption{Ablation study on the DNet. There are two curves: the red one 'D2CFR with DNet' and the blue one 'D2CFR w/o DNet' represent D2CFR with and without DNet respectively. X-axis represents the number of iterations.	For fig.a, Y-axis represents the winning from 'D2CFR with DNet' vs. 'D2CFR w/o DNet'. The higher, the better. For fig.b, the lower, the better.}
	\label{DNetresult}
\end{figure*}

\textbf{The effectiveness of DNet:} D2CFR takes DNet as the value network compared with the fully connected network in vanilla DeepCFR. 'D2CFR w/o DNet' can be regarded as the DeepCFR in this paper. The winning and policy loss are used to measure this improvement from the performance and the convergence of the policy network. Here the winning is obtained by head-to-head when the corresponding model at the same number of iteration. The results are shown in Fig.\ref{DNetresult}.

\begin{figure*}[t!]
	\centering
	\subfloat[Winning]{\includegraphics[width=0.48\linewidth,height=0.25\linewidth]{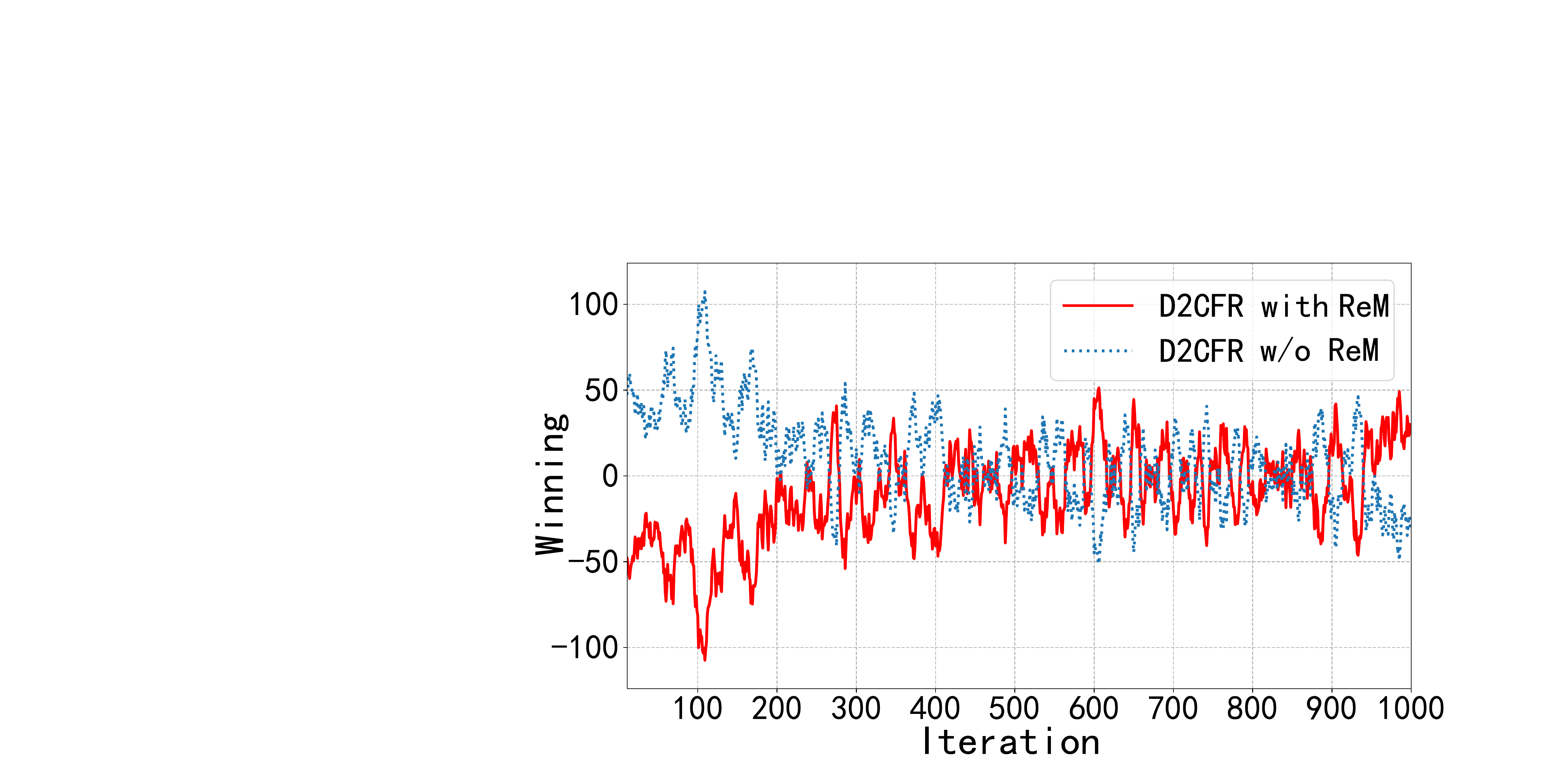}%
	}
	\hfil
	\subfloat[Value loss]{\includegraphics[width=0.48\linewidth,height=0.25\linewidth]{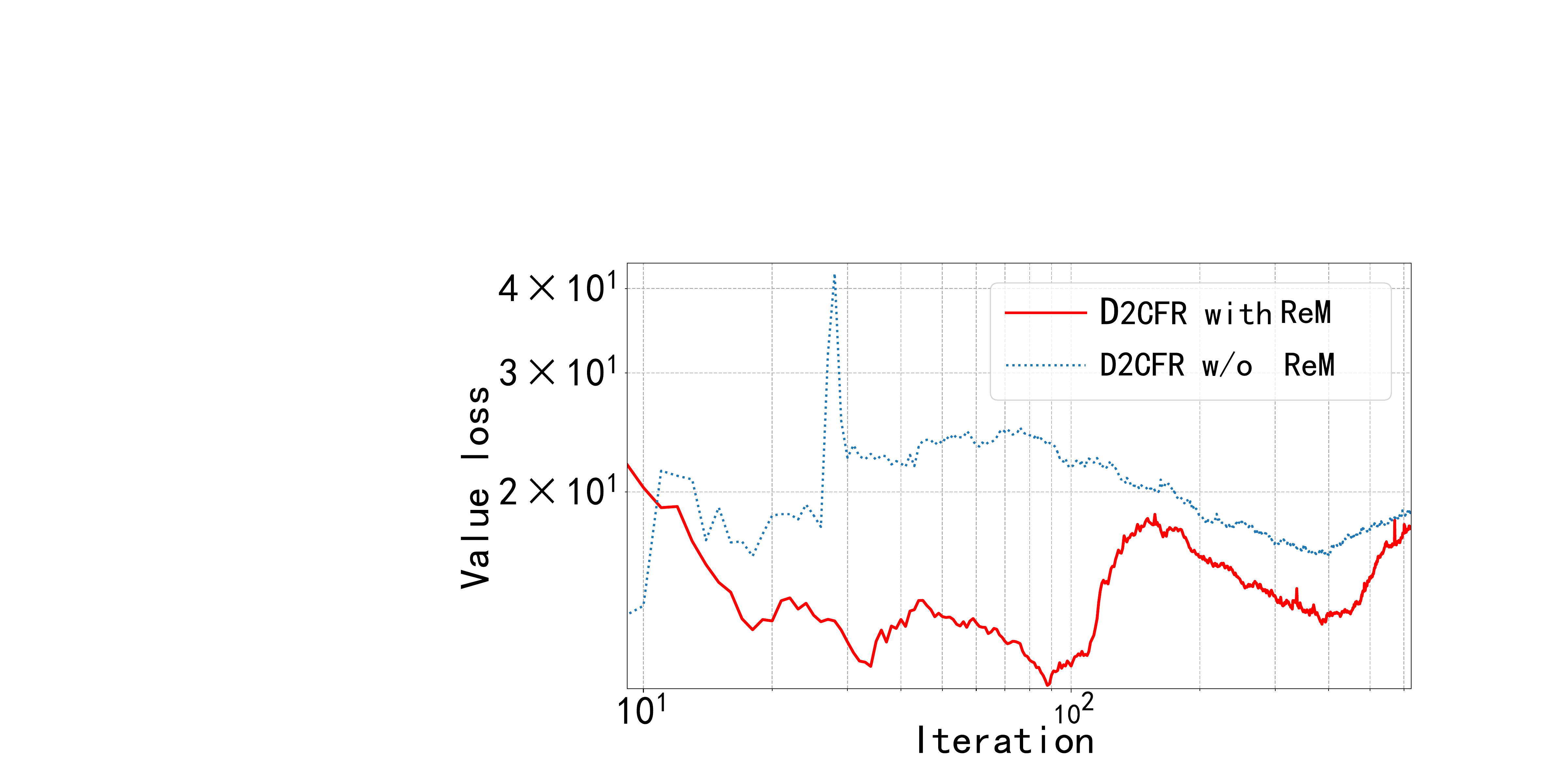}%
	}
	\caption{Ablation study on the rectification module. There are two curves: the red one 'D2CFR with ReM' and the blue one 'D2CFR w/o ReM' represent D2CFR with and without the rectification module respectively. X-axis represents the number of iterations. For fig.a, Y-axis represents the winning from 'D2CFR with ReM' vs. 'D2CFR w/o ReM'. The higher, the better. For fig.b, the lower, the better.}
	\label{mcresult}
\end{figure*}

It can be found that the DNet obviously improves the performance of D2CFR in Fig.\ref{DNetresult}(a). A positive winning means that our method is better than comparison methods. After $100th$ iterations, the 'D2CFR with DNet' is always better compared with the 'D2CFR w/o DNet' from the winning. Fig.\ref{DNetresult}(b) shows that the policy loss of the 'D2CFR with DNet' is far below than that of the 'D2CFR w/o DNet' all the time. And the gap of the two networks is widening after $120th$ iterations, which shows that our improvement is also very helpful to improve the convergence of the policy network.

\textbf{The effectiveness of rectification module:} The rectification module is the core of D2CFR when training whole neural networks, which is used to correct the inaccuracy of state estimation in the early stage, that is to reduce the approximation error of value network. Here the winning and the value loss are used to test the effectiveness of this component. In addition, 'ReM' is used to represent the rectification module in the following. 'D2CFR with ReM' and 'D2CFR w/o ReM' means D2CFR with and without the rectification module respectively.

It can be found that the rectification module obviously reduces the approximation error from Fig.\ref{mcresult}(b). The loss of the rectification module is always lower than that of the 'D2CFR w/o ReM' since the $12th$ iterations. And there is a clear gap from $20th$ to $500th$ iterations, which reflects that our rectification module is effective in reducing loss, especially in the early stage of iterations. Fig.\ref{mcresult}(a) shows that the winning is basically the same as that of 'D2CFR w/o ReM' after $200th$ iterations. Specifically, the winning has begun to increase from $100th$ iterations. This shows the performance of 'D2CFR with ReM' has not still decreased when the approximation error with the rectification module reduced.

\begin{figure*}
	\subfloat[Policy loss]{\includegraphics[width=0.5\linewidth,height=0.26\linewidth]{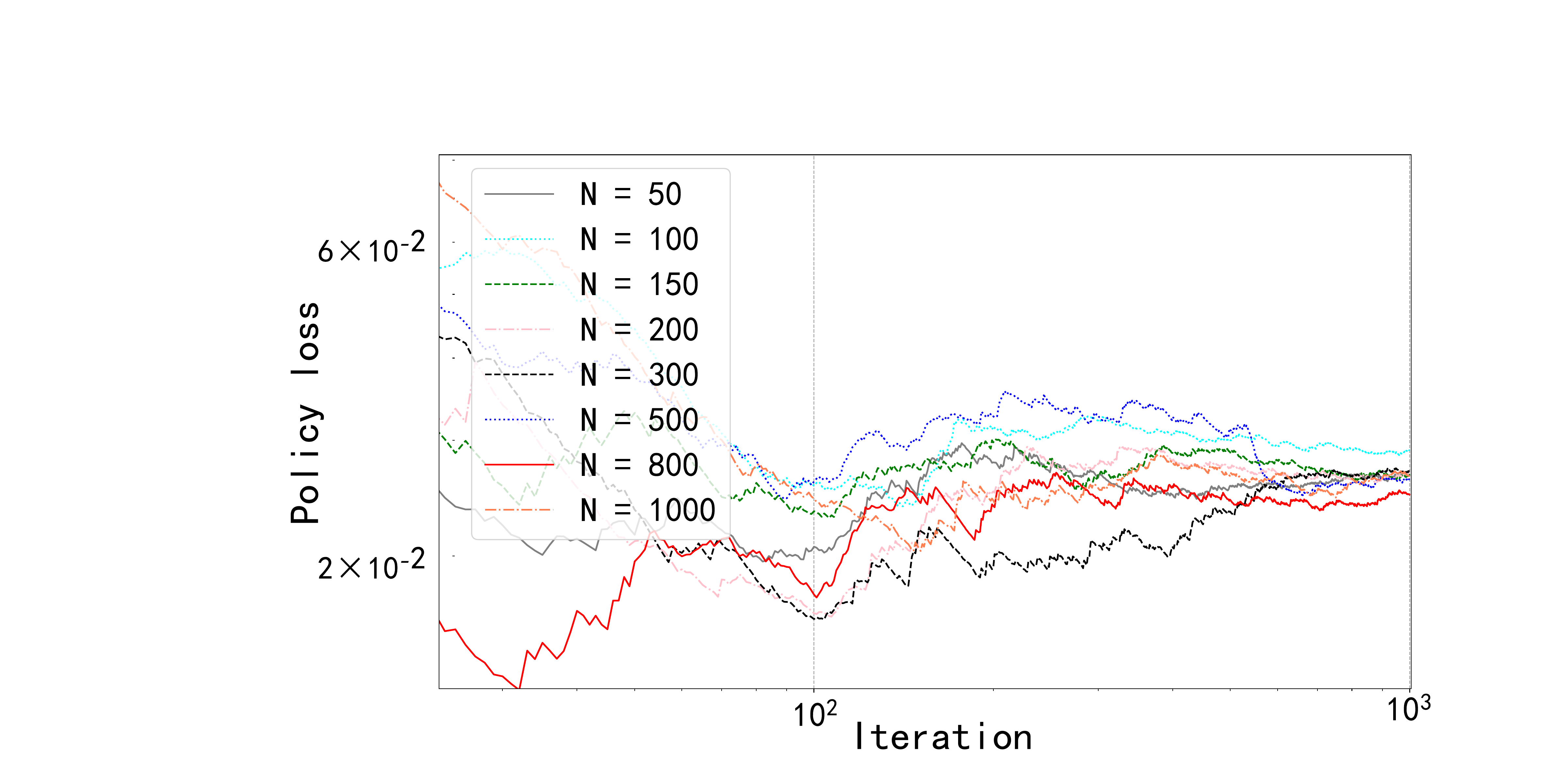}%
	}
	\hfil
	\subfloat[Exploitability]{\includegraphics[width=0.49\linewidth,height=0.26\linewidth]{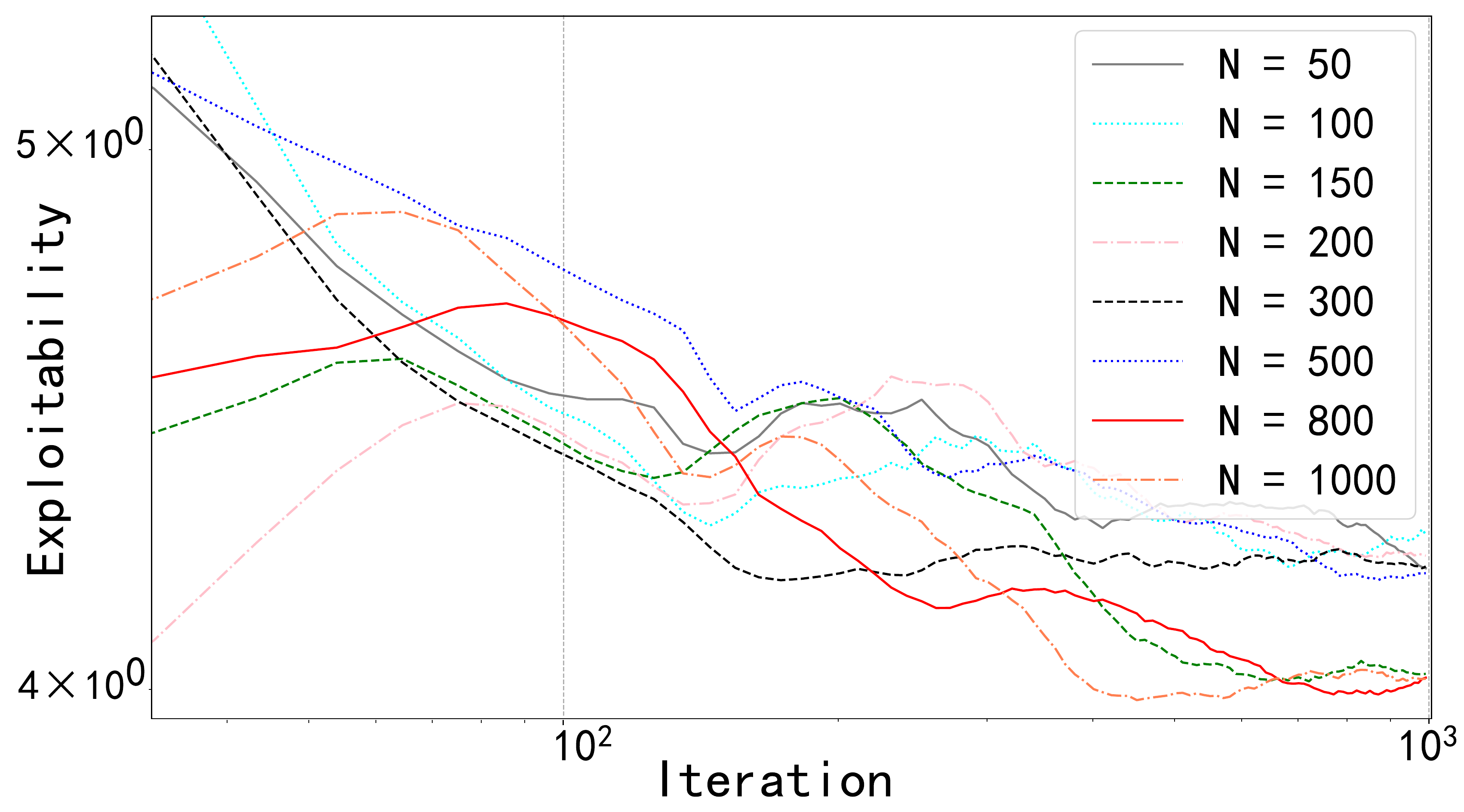}%
	}
	\caption{Ablation study on different times $N$ of MC. There are eight lines in the figure, which represent eight different setting values of MC times $N$. X-axis represents the number of iterations. For fig.a, Y-axis represents the exploitability, the lower, the better. For fig.b, Y-axis represents the policy losthe lower the loss, the better.}
	\label{mctimes}
\end{figure*}

Different setting values of $N$ in MC: Different setting values will have different effects on the MC simulation. Here we test eight different settings of $N$, $N=50,100,150,200,300,500,800,1000$. The policy loss and exploitability are used to evaluate its performance for selecting an optimal value of $N$. The experiment results are shown in Fig.\ref{mctimes}.

It can be found that the exploitability of $N=800$ and $N=1000$ are lower than that of other settings from Fig.\ref{mctimes}(b). And the exploitability of $N=800$ is better than that of $N=1000$ except $350th\sim680th$ iterations.  Fig.\ref{mctimes}(a) shows that $N=300$ and $N=800$ are better than that of others in terms of the policy loss. And the policy loss of $N=800$ is lower than that of $N=300$ except $150th\sim550th$ iterations. Therefore, considering these three aspects, we chose the times $N=800$ as the final setting in our experiment.

\section{Conclusion}
In this paper, we present an improved version D2CFR based on the DeepCFR, which can compute approximate Nash equilibrium in two-player imperfect-information games. The D2CFR constructs a DNet as the value network that decouples the state value estimation and action-state value estimation, which can provide accurate state value by estimating it explicitly. Moreover, a rectification module based on Monte Carlo simulation is designed, which can further rectify the error estimation of states in the early stage. Extensive experimental results show that the improvement of our D2CFR is effective, and the D2CFR outperforms other state-of-the-art methods on test games. In the future, better fine-tuned network architecture will be helpful to improve the performance of D2CFR. In addition, it would be interesting to expend D2CFR to larger and more complex games than poker games. 

\section*{Acknowledgments}
This research was funded by key fields R\&D project of Guangdong Province (No.2020B0101380001), National Natural Science Founda-tion of China (No.61902093), Natural Science Foundation of Guang-dong (No.2020A1515010652), Shenzhen Foundational Research Fund-ing Under Grant (No.20200805173048001), PINGAN-HITsz Intelli-gence Finance Research Center, Ricoh-HITsz Joint Research Center, GBase-HITsz Joint Research Center.

 
\bibliographystyle{IEEEtran}
\bibliography{conferenceforCFRNN}

\begin{thebibliography}{10}
\providecommand{\url}[1]{#1}
\csname url@samestyle\endcsname
\providecommand{\newblock}{\relax}
\providecommand{\bibinfo}[2]{#2}
\providecommand{\BIBentrySTDinterwordspacing}{\spaceskip=0pt\relax}
\providecommand{\BIBentryALTinterwordstretchfactor}{4}
\providecommand{\BIBentryALTinterwordspacing}{\spaceskip=\fontdimen2\font plus
\BIBentryALTinterwordstretchfactor\fontdimen3\font minus
  \fontdimen4\font\relax}
\providecommand{\BIBforeignlanguage}[2]{{%
\expandafter\ifx\csname l@#1\endcsname\relax
\typeout{** WARNING: IEEEtran.bst: No hyphenation pattern has been}%
\typeout{** loaded for the language `#1'. Using the pattern for}%
\typeout{** the default language instead.}%
\else
\language=\csname l@#1\endcsname
\fi
#2}}
\providecommand{\BIBdecl}{\relax}
\BIBdecl

\bibitem{samuel1959some}
A.~L. Samuel, ``Some studies in machine learning using the game of checkers,''
  \emph{IBM Journal of research and development}, vol.~3, no.~3, pp. 210--229,
  1959.

\bibitem{myerson1997game}
R.~B. Myerson, \emph{Game theory: analysis of conflict}.\hskip 1em plus 0.5em
  minus 0.4em\relax Cambridge, 1997.

\bibitem{DengN19}
Z.~Deng and X.~Nian, ``Distributed generalized nash equilibrium seeking
  algorithm design for aggregative games over weight-balanced digraphs,''
  \emph{{IEEE} Trans. Neural Networks Learn. Syst.}, vol.~30, no.~3, pp.
  695--706, 2019.

\bibitem{ZhangYYL19}
P.~Zhang, Y.~Yuan, H.~Yang, and H.~Liu, ``Near-nash equilibrium control
  strategy for discrete-time nonlinear systems with round-robin protocol,''
  \emph{{IEEE} Trans. Neural Networks Learn. Syst.}, vol.~30, no.~8, pp.
  2478--2492, 2019.

\bibitem{LiQMZK21}
M.~Li, J.~Qin, Q.~Ma, W.~X. Zheng, and Y.~Kang, ``Hierarchical optimal
  synchronization for linear systems via reinforcement learning: {A}
  stackelberg-nash game perspective,'' \emph{{IEEE} Trans. Neural Networks
  Learn. Syst.}, vol.~32, no.~4, pp. 1600--1611, 2021.

\bibitem{osborne1994course}
M.~J. Osborne and A.~Rubinstein, \emph{A course in game theory}.\hskip 1em plus
  0.5em minus 0.4em\relax MIT press, 1994.

\bibitem{nash1951non}
J.~Nash, ``Non-cooperative games,'' \emph{Annals of mathematics}, pp. 286--295,
  1951.

\bibitem{zinkevich2008regret}
M.~Zinkevich, M.~Johanson, M.~Bowling, and C.~Piccione, ``Regret minimization
  in games with incomplete information,'' in \emph{Advances in neural
  information processing systems}, 2008, pp. 1729--1736.

\bibitem{bowling2015heads}
M.~Bowling, N.~Burch, M.~Johanson, and O.~Tammelin, ``Heads-up limit hold’em
  poker is solved,'' \emph{Science}, vol. 347, no. 6218, pp. 145--149, 2015.

\bibitem{moravvcik2017deepstack}
M.~Morav{\v{c}}{\'\i}k, M.~Schmid, N.~Burch, V.~Lis{\`y}, D.~Morrill, N.~Bard,
  T.~Davis, K.~Waugh, M.~Johanson, and M.~Bowling, ``Deepstack: Expert-level
  artificial intelligence in heads-up no-limit poker,'' \emph{Science}, vol.
  356, no. 6337, pp. 508--513, 2017.

\bibitem{Brown2017Superhuman}
N.~Brown and T.~Sandholm, ``Superhuman ai for heads-up no-limit poker: Libratus
  beats top professionals.'' \emph{Science}, vol. 359, no. 6374, p. 1733, 2017.

\bibitem{schmid2019variance}
M.~Schmid, N.~Burch, M.~Lanctot, M.~Moravcik, R.~Kadlec, and M.~Bowling,
  ``Variance reduction in monte carlo counterfactual regret minimization
  (vr-mccfr) for extensive form games using baselines,'' in \emph{Proceedings
  of the AAAI Conference on Artificial Intelligence}, vol.~33, 2019, pp.
  2157--2164.

\bibitem{brown2019superhuman}
N.~Brown and T.~Sandholm, ``Superhuman ai for multiplayer poker,''
  \emph{Science}, vol. 365, no. 6456, pp. 885--890, 2019.

\bibitem{brown2019deep}
N.~Brown, A.~Lerer, S.~Gross, and T.~Sandholm, ``Deep counterfactual regret
  minimization,'' in \emph{International Conference on Machine Learning}, 2019,
  pp. 793--802.

\bibitem{li2018double}
H.~Li, K.~Hu, Z.~Ge, T.~Jiang, Y.~Qi, and L.~Song, ``Double neural
  counterfactual regret minimization,'' \emph{arXiv preprint arXiv:1812.10607},
  2018.

\bibitem{steinberger2019single}
E.~Steinberger, ``Single deep counterfactual regret minimization,'' \emph{arXiv
  preprint arXiv:1901.07621}, 2019.

\bibitem{heinrich2015fictitious}
J.~Heinrich, M.~Lanctot, and D.~Silver, ``Fictitious self-play in
  extensive-form games,'' in \emph{International conference on machine
  learning}.\hskip 1em plus 0.5em minus 0.4em\relax PMLR, 2015, pp. 805--813.

\bibitem{2016Deep}
J.~Heinrich and D.~Silver, ``Deep reinforcement learning from self-play in
  imperfect-information games,'' \emph{arXiv:1603.01121}, 2016.

\bibitem{rowley1998neural}
H.~A. Rowley, S.~Baluja, and T.~Kanade, ``Neural network-based face
  detection,'' \emph{IEEE Transactions on pattern analysis and machine
  intelligence}, vol.~20, no.~1, pp. 23--38, 1998.

\bibitem{Shapley1996Fictitious}
M.~L.~S. Shapley, ``Fictitious play property for games with identical
  interests,'' \emph{Journal of Economic Theory}, 1996.

\bibitem{lambert2005fictitious}
T.~J. Lambert~Iii, M.~A. Epelman, and R.~L. Smith, ``A fictitious play approach
  to large-scale optimization,'' \emph{Operations Research}, vol.~53, no.~3,
  pp. 477--489, 2005.

\bibitem{shamma2005dynamic}
J.~S. Shamma and G.~Arslan, ``Dynamic fictitious play, dynamic gradient play,
  and distributed convergence to nash equilibria,'' \emph{IEEE Transactions on
  Automatic Control}, vol.~50, no.~3, pp. 312--327, 2005.

\bibitem{Martin1994}
M.~J. Osborne and A.~Rubinstein, \emph{A Course in Game Theory}.\hskip 1em plus
  0.5em minus 0.4em\relax The MIT Press, 1994.

\bibitem{foster1999regret}
D.~P. Foster and R.~Vohra, ``Regret in the on-line decision problem,''
  \emph{Games and Economic Behavior}, vol.~29, no. 1-2, pp. 7--35, 1999.

\bibitem{lanctot2009monte}
M.~Lanctot, K.~Waugh, M.~Zinkevich, and M.~Bowling, ``Monte carlo sampling for
  regret minimization in extensive games,'' in \emph{Advances in neural
  information processing systems}, 2009, pp. 1078--1086.

\bibitem{shi2000abstraction}
J.~Shi and M.~L. Littman, ``Abstraction methods for game theoretic poker,'' in
  \emph{International Conference on Computers and Games}.\hskip 1em plus 0.5em
  minus 0.4em\relax Springer, 2000, pp. 333--345.

\bibitem{gilpin2007potential}
A.~Gilpin, T.~Sandholm, and T.~B. S{\o}rensen, ``Potential-aware automated
  abstraction of sequential games, and holistic equilibrium analysis of texas
  hold'em poker,'' in \emph{Proceedings of the National Conference on
  Artificial Intelligence}, vol.~22, no.~1.\hskip 1em plus 0.5em minus
  0.4em\relax Menlo Park, CA; Cambridge, MA; London; AAAI Press; MIT Press;
  1999, 2007, p.~50.

\bibitem{gilpin2006competitive}
A.~Gilpin and T.~Sandholm, ``A competitive texas hold'em poker player via
  automated abstraction and real-time equilibrium computation,'' in
  \emph{AAAI}, 2006, pp. 1007--1013.

\bibitem{gilpin2007better}
------, ``Better automated abstraction techniques for imperfect information
  games, with application to texas hold'em poker,'' in \emph{Proceedings of the
  6th international joint conference on Autonomous agents and multiagent
  systems}, 2007, pp. 1--8.

\bibitem{ganzfried2013action}
S.~Ganzfried and T.~Sandholm, ``Action translation in extensive-form games with
  large action spaces: Axioms, paradoxes, and the pseudo-harmonic mapping,'' in
  \emph{Workshops at the Twenty-Seventh AAAI Conference on Artificial
  Intelligence}, 2013.

\bibitem{zio2013monte}
E.~Zio, ``Monte carlo simulation: The method,'' in \emph{The Monte Carlo
  simulation method for system reliability and risk analysis}.\hskip 1em plus
  0.5em minus 0.4em\relax Springer, 2013, pp. 19--58.

\bibitem{2012A}
C.~B. Browne, E.~Powley, D.~Whitehouse, S.~M. Lucas, P.~I. Cowling,
  P.~Rohlfshagen, S.~Tavener, D.~Perez, S.~Samothrakis, and S.~Colton, ``A
  survey of monte carlo tree search methods,'' \emph{IEEE Transactions on
  Computational Intelligence \& Ai in Games}, vol.~4, no.~1, pp. 1--43, 2012.

\bibitem{bu2018estimation}
Y.~Bu, S.~Zou, Y.~Liang, and V.~V. Veeravalli, ``Estimation of kl divergence:
  Optimal minimax rate,'' \emph{IEEE Transactions on Information Theory},
  vol.~64, no.~4, pp. 2648--2674, 2018.

\bibitem{southey2005bayes}
F.~Southey, M.~Bowling, B.~Larson, C.~Piccione, N.~Burch, D.~Billings, and
  C.~Rayner, ``Bayes' bluff: opponent modelling in poker,'' in
  \emph{Proceedings of the Twenty-First Conference on Uncertainty in Artificial
  Intelligence}, 2005, pp. 550--558.

\bibitem{lanctot2019openspiel}
M.~Lanctot, E.~Lockhart, J.-B. Lespiau, V.~Zambaldi, S.~Upadhyay,
  J.~P{\'e}rolat, S.~Srinivasan, F.~Timbers, K.~Tuyls, S.~Omidshafiei
  \emph{et~al.}, ``Openspiel: A framework for reinforcement learning in
  games,'' \emph{arXiv preprint arXiv:1908.09453}, 2019.

\bibitem{kingma2014adam}
D.~P. Kingma and J.~Ba, ``Adam: A method for stochastic optimization,''
  \emph{arXiv preprint arXiv:1412.6980}, 2014.

\end{thebibliography}

\vfill

\end{document}